\documentclass[letterpaper]{article}

\usepackage{aaai23}
\nocopyright
\usepackage{natbib}
\usepackage{times}
\usepackage{helvet}
\usepackage{courier}
\usepackage{xcolor}
\usepackage{mathtools}
\usepackage{algorithm}
\usepackage{algpseudocode}
\usepackage{graphicx} 
\usepackage[hyphens]{url}
\usepackage{float}
\usepackage{listings}
\usepackage{booktabs}
\usepackage{mathdots}
\usepackage{soul}

\usepackage[utf8]{inputenc}
\usepackage{xspace}
\usepackage{subcaption}
\usepackage{tikz}
\usepackage{cleveref}

\usepackage{listings}
\usepackage{amsthm,amsmath,amsfonts}
\theoremstyle{definition}

\theoremstyle{thmstyleone}%

\usetikzlibrary{matrix, positioning}

\newcommand{\savilerow}{{Savile Row}\xspace}

\newcommand{\zap}[1]{} 

\definecolor{mygreen}{rgb}{0,0.6,0}

\newcommand{\cR}{\colorbox{red!30}{R}}
\newcommand{\cG}{\colorbox{green!30}{G}}

\newcommand{\drawGrid}[2]{%
\draw[xstep=0.5cm,ystep=0.5cm,color=gray] (0,0) grid (#1,#2);%
}
\newcommand{\matrixGameHead}{%
\matrix (game) [matrix of nodes, inner sep=0pt, anchor=south west, nodes={inner sep=0pt,text width=.5cm,align=center,minimum height=.5cm}]
}
\newcommand{\drawGround}[2]{%
\draw[line width=0.6pt] (game.south) (#1,0) -- (#2,0);%
}

\usepackage{listings}
\lstdefinelanguage{PDDL}
{
  sensitive=false,    
  morecomment=[l]{;}, 
  alsoletter={:,-},   
  morekeywords={
    define,domain,problem,not,and,or,when,forall,exists,either,
    :domain,:requirements,:types,:objects,:constants,
    :predicates,:action,:parameters,:precondition,:effect,
    :fluents,:primary-effect,:side-effect,:init,:goal,
    :strips,:adl,:equality,:typing,:conditional-effects,
    :negative-preconditions,:disjunctive-preconditions,
    :existential-preconditions,:universal-preconditions,:quantified-preconditions,
    :functions,assign,increase,decrease,scale-up,scale-down,
    :metric,minimize,maximize,
    :durative-actions,:duration-inequalities,:continuous-effects,
    :durative-action,:duration,:condition
  }
}

\author {
    Joan Espasa,\textsuperscript{\rm 1}
    Ian Miguel, \textsuperscript{\rm 1}
    Peter Nightingale,\textsuperscript{\rm 2}
    András Z. Salamon,\textsuperscript{\rm 1}
    Mateu Villaret\textsuperscript{\rm 3}
}
\affiliations {
    \textsuperscript{\rm 1}School of Computer Science, University of St Andrews, UK\\
    
    \textsuperscript{\rm 2}Department of Computer Science, University of York, UK\\

    \textsuperscript{\rm 3}Departament d'Inform\`atica, Matem\`atica Aplicada i Estad\'istica, Universitat de Girona, Spain\\
    
    \{jea20,ijm\}@st-andrews.ac.uk, 
    peter.nightingale@york.ac.uk,
    Andras.Salamon@st-andrews.ac.uk,
    mateu.villaret@udg.edu
}

\begin{document}

\title{Challenges in Modelling and Solving Plotting with PDDL
\thanks{A paper extending this work and~\cite{espasa2022plotting} has been submitted for journal publication.}
}

\maketitle

\begin{abstract}
\begin{quote}
We study a planning problem based on Plotting, a tile-matching puzzle video game published by Taito in 1989.
The objective of this game is to remove a target number of coloured blocks from a grid by sequentially shooting blocks into the grid. Plotting features complex transitions after every shot: various blocks are affected directly, while others can be indirectly affected by gravity. 
We highlight the challenges of modelling Plotting with PDDL and of solving it with a grounding-based state-of-the-art planner.
\end{quote}
\end{abstract}

\section{Introduction}
\begin{figure}[t!] 
\centering
\includegraphics[scale=0.18]{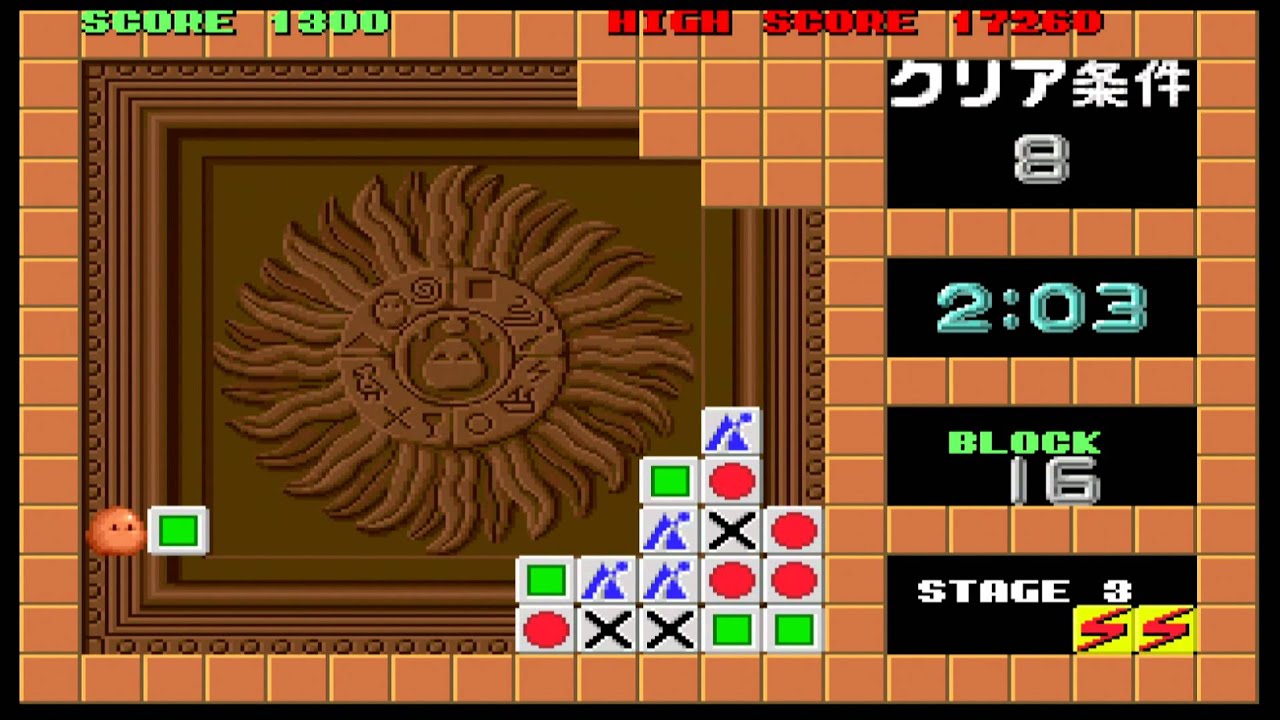}
\caption{Plotting (Taito, 1989). The avatar is seen on the left, holding a green block. The objective is to reduce the number of blocks in the middle pile. In this particular case there are 16 left (see center-right of the image), and the goal is 8 or less (see top-right of image).}
\label{fig:screenshot}
\end{figure}

We consider finding optimal solutions for a discrete time and space puzzle, \emph{Plotting}, a puzzle video game published by Taito in 1989 and ported to many platforms. The objective is to reduce a given grid of coloured blocks to a goal number or fewer (Figure \ref{fig:screenshot}). This is achieved by the avatar character repeatedly shooting the block it holds into the grid. The game is also known as \emph{Flipull} in Japan as well as in versions for the Famicom and Game Boy.

Plotting is naturally characterised as a planning problem~\cite{csplib:prob088}, aiming to find a sequence of firing positions such that enough blocks are removed to beat the scenario objective. The complexity of state transitions after every shot makes this problem interesting: some blocks are affected directly, while others can be indirectly affected by gravity, as explained in the next section. Modelling the game dynamics in PDDL~\cite{pddl} is difficult, as we will demonstrate.

The resulting complexity of the model severely hinders the ability of current planning systems to produce a valid plan. 
Most state-of-the-art AI planners rely on grounding, instantiating every action schema for all meaningful combinations of parameters.
As we will show, the complexity of Plotting is too much for this grounding process. Problems with grounding are now attracting attention in the planning community~\cite{matloob2016:exploring, powerlifted}, with suggestions to avoid grounding {\em lifted} representations as far as possible. A lifted representation succinctly defines actions by grouping them with their preconditions and effects using action schemas with parameters.

Constraint modelling languages can be used to express planning problems~\cite{bartak2010constraint,related2,related1,modref20,cplan}. These languages are more expressive than PDDL and permit a succinct lifted representation of Plotting, providing access to lifted solving approaches that don't need exhaustive grounding. However, they are not a panacea and require significant human effort to be put into modelling.
In this work we describe Plotting and provide a working model, an instance generator and a set of benchmark instances. We also highlight the challenges of modelling and solving Plotting with PDDL.

\section{\label{sec:Plotting}Plotting}

Plotting is played by one agent with full information of the game state, and the effects of each action are deterministic. This situation is common in puzzle-style video games, and similar to pen and paper puzzles~\citep{puzzles}, some variants of patience like Black Hole \cite{gent2007search}, and board games such as peg solitaire~\cite{pegsolitaire} or the knight's tour~\cite{knightstour}. The objective in Plotting is to reduce a given grid of coloured blocks down to a goal number or fewer. This is achieved by the avatar character shooting the block it holds into the grid, either horizontally directly into the grid, or by shooting at the wall blocks above the grid, and bouncing down vertically onto the grid. Note that we consider the topmost row as the first row and the leftmost column as the first column.
When shooting a block, if it hits a wall as it is travelling horizontally, it falls vertically downwards. In a typical level, additional walls are arranged to facilitate hitting the blocks from above. If the block falls onto the floor, it rebounds into the avatar's hand.

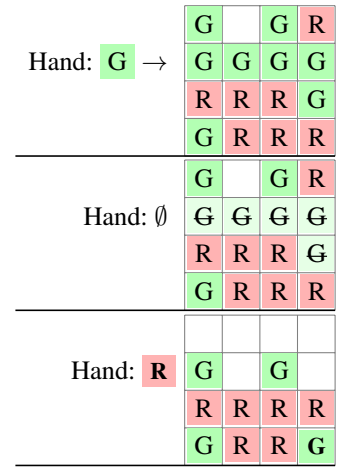
\begin{figure}[ht]
\centering
  \begin{subfigure}{0.31\textwidth}
    \centering
    \begin{tikzpicture}
    \drawGrid{2}{2}
    \matrixGameHead
    {
    \cG &   & \cG & \cR \\
    \cG & \cG & \cG & \cG \\
    \colorbox{red!30}R & \cR & \cR & \cG \\
    \cG & \cR & \cR & \cR \\
    };
    \drawGround{-2.25}{2}
    \node (hand)[left=0.1cm of game.west, yshift=0.25cm] {Hand: \colorbox{green!30}{G}\ $\to$};
    \end{tikzpicture}
  \end{subfigure}
    \begin{subfigure}{0.31\textwidth}
  \centering
    \begin{tikzpicture}
    \drawGrid{2}{2}
    \matrixGameHead
    {
    \cG &   & \cG & \cR \\
    \colorbox{green!10}{\st{G}} & \colorbox{green!10}{\st{G}} & \colorbox{green!10}{\st{G}} & \colorbox{green!10}{\st{G}} \\
    \cR & \cR & \cR & \colorbox{green!10}{\st{G}} \\
    \cG & \cR & \cR & \cR \\
    };
    \drawGround{-2.25}{2}
    \node (hand)[left=0.1cm of game.west, yshift=0.25cm] {Hand:  $\emptyset$ }; 
    \end{tikzpicture}
  \end{subfigure}
  \begin{subfigure}{0.31\textwidth}
  \centering
    \begin{tikzpicture}
    \drawGrid{2}{2}
    \matrixGameHead
    {
      &   &   & ~ \\
    \cG &   & \cG &   \\
    \cR & \cR & \cR & \cR \\
    \cG & \cR & \cR & \textbf{\small\cG} \\
    };
    \drawGround{-2.25}{2}
    \node (hand)[left=0.0cm of game.west, yshift=0.25cm] {Hand: \colorbox{red!30}{\small\textbf{R}}};
    \end{tikzpicture}
  \end{subfigure}
  \caption{A diagram of a shot where the firing block reaches the end and goes downwards. The top right red block has to fall a variable number of positions (two in this case), depending on the state of the board and the colour of the shot.}
  \label{fig:rowcolshot}
\end{figure}

The rules for a shot block $S$ colliding with a block $B$ in the grid are a bit more complex:
\begin{itemize}

\item  If the first block $S$ hits is of a different type from itself, $S$ rebounds into the avatar’s hand and the grid is unchanged: this is a null move.

\item  If \emph{S} and \emph{B} are of the same type, \emph{B} is consumed and \emph{S} continues to travel in the same direction. All blocks above \emph{B} fall one grid cell each.

\item  If \emph{S}, having already consumed a block of the same type, hits a block \emph{B} of a different type, then \emph{S} replaces \emph{B}, and \emph{B} rebounds into the avatar’s hand.

\end{itemize}
A complex shot is depicted in Figure~\ref{fig:rowcolshot}, where a green block consumes an entire row of the grid, hits the wall, and continues to consume blocks as it falls until it finds a block of a different colour (red). Finally, the green block replaces the final red block, which rebounds to the avatar's hand. Blocks above the consumed green blocks fall.
If, after making a shot, the block that rebounds into the avatar’s hand is such that there is now no possible shot that can further reduce the grid, we reach a dead end and the block in the avatar’s hand is transformed into a wildcard block, which transforms into the same type as the first block it hits. Each level also begins with the avatar holding a wildcard block.
In our models we consider the task of finding a solution while avoiding dead ends, since each dead end causes the loss of one of the player's lives.
 
Plotting's initial state is the given grid, and there are usually multiple goal states where the grid is sufficiently reduced to meet the target. In the model we abstract the avatar's movement to consider the key decisions: the rows or columns chosen at which to shoot the held block. Therefore, the sequence of actions to get us from the initial to the goal state is comprised of individual shots at the grid, either horizontally or vertically.

\section{Limitations of Planning Approaches}\label{sec:limitations}

Tools to solve problems such as Plotting should ideally support natural ways of expressing elements such as matrices to represent the state of play, a way to index the entries in such matrices, and a representation of the states of the blocks.
PDDL 2.1~\cite{pddl21} added support for numeric and temporal features, extending the expressivity of the language. Still, such an extension is insufficient for efficiently modelling and solving Plotting. \citet{pddl21} states:

\begin{quote}
    Numeric expressions are not allowed to appear as terms in the language (that is, as arguments to predicates or values of action parameters) \dots Functions in PDDL2.1 are restricted to be of type $\mathit{Object}_n \rightarrow \mathbb{R}$, for the (finite) collection of objects in a planning instance, $\mathit{Object}$ and finite function arity $n$. 
\end{quote}

In other words, no action, predicate or function can have a number as a parameter. Sadly, these severe limitations render this PDDL extension useless for our needs. Note that an essential construct in the preconditions and effects of the actions would be the use of arithmetic to deal with indices of rows and columns that actions should have as parameters. 
For example,  when we remove a block in a given {\tt row} and {\tt col}, if there was a block above it, this block would fall and we would need to refer to its colour. 
Unfortunately, since {\tt row} cannot be a numeric parameter in PDDL, we are forced to use quantifiers to be able to refer to the ``block that is above it'' (i.e.~its row is equal to {\tt row+1}). Therefore, as we will see in the next section, we are forced to define predicates to simulate some basic arithmetic operations on indices.

Functional STRIPS~\cite{fstrips} or Planning Modulo Theories~\cite{pmt} would alleviate the expressivity problems faced with Plotting. On one hand, with Functional STRIPS extensions such as those of~\citet{fstrips2} we would be able to both simulate matrices thanks to proper support for functions in the language, and to operate on their indices thanks to the arithmetic support. On the other hand, the Planning Modulo Theories paradigm could support a theory with both matrices and arithmetic.

Unfortunately, these approaches have either not been released, or are not actively maintained, and we have been unable to use these off the shelf.
In particular, the FS planner~\cite{fstrips2} has dependencies on software which we have not been able to find and
the Planning Modulo Theories planner of~\citet{pmt} was not released. Therefore, it would require a significant engineering effort to either reproduce or re-engineer them. Considering available and well supported planners, we are limited to using classical planners.

\section{Modelling Plotting in PDDL}\label{sec:pddl}

We now provide fragments of the model to illustrate the main drawbacks of PDDL for modelling Plotting. The game board is abstracted as a grid of coloured cells. The colour of each cell is the colour of the block it contains, or \texttt{null} if empty. Therefore, the state is the colour of each cell and the colour of the block in the avatar's hand.
To parameterise the actions and the predicates defining the state, we use two types of objects: \texttt{colour} and \texttt{number}, where \texttt{number} is the name of a type used to manually encode the basic required numerical properties. The predicate \texttt{hand} has one colour parameter, and encodes if the avatar has a block of the given colour. Given parameters \texttt{row}, \texttt{col} and \texttt{c}, the  \texttt{coloured} predicate expresses if the block in that row and column has the given colour.
\begin{lstlisting}[escapechar=|, language=PDDL]
(hand ?c - colour)
(coloured ?row ?col - number ?c - colour)
\end{lstlisting}
Auxiliary predicates such as \texttt{islastcolumn} or \texttt{isbottomrow} are added both for clarity and to reduce the use of quantifiers and so the burden on the planner's preprocessor. 
\begin{lstlisting}[escapechar=|, language=PDDL]
(isfirstcolumn ?n - number)
(islastcolumn ?n - number)
(istoprow ?n - number)
(isbottomrow ?n - number)
\end{lstlisting}
Moreover, we need to encode some integer relations as Boolean predicates:
\begin{lstlisting}[escapechar=|, language=PDDL]
(succ ?p1 ?p2 - number)         ; p1 is successor of p2
(lt ?p1 ?p2 - number)           ; p1 is less than p2
(distance ?p1 ?p2 ?p3 - number) ; p3 is p2 - p1
\end{lstlisting}
These predicates must be defined in each instance file, along with the specific scenario information. For instance, when dealing with a $5\times 5$ board we need to state {\tt succ} for every pair of successive numbers between 1 and 5, and {\tt lt} and {\tt distance} for every pair of two numbers $(p_1,p_2)$ between 1 and 5 such that $p_1<p_2$. 
\lstset{
  basicstyle=\scriptsize\ttfamily,
  columns=fullflexible, keepspaces=true,
  numbers=left, tabsize=2, stepnumber=1, language=PDDL }
\begin{figure}[t!]
\begin{lstlisting}[escapechar=|]
(:action shoot-partial-row
  ;; ?r - what row we are shooting at
  ;; ?t - the end cell where the shot ends
  ;; ?c - the colour we are removing
  :parameters (?r - number ?t - number ?c - colour)
  :precondition (and
    ;; ?col is successor of ?t coloured differently to ?c
    (exists (?col - number)|\label{line:p1}|
      (and (succ ?col ?t) 
           (not (coloured ?r ?col ?c))
           (not (coloured ?r ?col null))))
    ...
    ;; all blocks up to ?t are either colour ?c or null
    (forall (?col - number) |\label{line:p2}|
      (or  (lt ?t ?col) 
           (and (= ?col ?t) (coloured ?r ?t ?c))
           (or (coloured ?r ?col ?c)
               (coloured ?r ?col null)))))
  :effect (and
    ;; Change hand colour
    ;; next cell that we cannot remove gets hand colour
    (forall (?nextcolumn - number ?nextcolour - colour)|\label{line:nextcolorquant}|
      (when |\label{line:when}|
        (and (succ ?nextcolumn ?t)
             (coloured ?r ?nextcolumn ?nextcolour))|\label{line:nextcolourwhen}|
        (and (not (coloured ?r ?nextcolumn ?nextcolour))
             (coloured ?r ?nextcolumn ?c)
             (hand ?nextcolour) |\label{line:hand1}|
             (not (hand ?c))))) |\label{line:hand2}|
    ;; Move everything downwards.
    ;; 2 cases: base case (top row), general case (rest)
    (forall (?currentrow ?nextrow ?currentcol - number)|\label{line:p3}|
      (and ;; the general case: any row except the top 
        (forall (?currentcolor ?nextcolor - colour)
          (when
            (and
              (lt ?currentrow ?r)
              (succ ?nextrow ?currentrow)
              (or (lt ?currentcol ?t)
                  (= ?currentcol ?t))
              ;; ensure cells have the pertaining colours
              (coloured ?currentrow ?currentcol ?currentcolor)
              (coloured ?nextrow ?currentcol ?nextcolor)
              ;; avoid a contradiction:
              (not (= ?currentcolor ?nextcolor)))
            (and
              (not (coloured ?nextrow ?currentcol ?nextcolor))
              (coloured ?nextrow ?currentcol ?currentcolor)
            ))))))); Then, case of firing on the top row.
        ...))
\end{lstlisting}
\caption{Fragment of the \emph{shoot-partial-row} action.}
\label{fig:pddl-code}
\end{figure}
\lstset{
  basicstyle=\scriptsize\ttfamily,
  columns=fullflexible, keepspaces=true,
  tabsize=2, stepnumber=1, numbers=none
}
Figure~\ref{fig:pddl-code} is an excerpt of the action consisting of partially removing blocks of colour {\tt ?c} in row {\tt ?r} until column {\tt ?t}, not reaching the last column. One of the principal difficulties is in identifying successors and predecessors of particular rows or columns (e.g. Lines~\ref{line:p1}, \ref{line:p2}, \ref{line:nextcolorquant}, \ref{line:p3}), which could have been eased by support for arithmetic on parameters.

The lack of support for multi-valued variables makes the encoding of some transitions difficult. For example, when changing the colour held by the avatar we must state: {\em remove previous colour in the hand and set the new colour} (lines \ref{line:hand1}-\ref{line:hand2}). Multi-valued variables would make this change straightforward. Due to the lack of support for function symbols in the considered PDDL fragment, we must also employ quantification to name specific objects. For instance, the column of the cell next to {\tt ?t} (\texttt{?nextcolumn}) and its colour (\texttt{?nextcolour}) have to be discovered. This quantification is introduced in line \ref{line:nextcolorquant}, and the values of \texttt{?nextcolumn} and \texttt{?nextcolour} are discovered in lines~\ref{line:when}-\ref{line:nextcolourwhen} as a condition for the effect to take place.

If we could use function symbols and arithmetic, we could remove variables \texttt{?nextcolumn} and \texttt{?nextcolour}, changing the \texttt{coloured} symbol to a function that, given a row and column, maps to the colour in that cell. Overall, lines \ref{line:nextcolorquant}-\ref{line:hand2} could theoretically be simplified to:
%
\begin{lstlisting}[escapechar=|, language=PDDL]
(assign (hand (coloured ?r (?t + 1))))
(assign (coloured ?r (?t + 1)) ?c)
\end{lstlisting}
Unfortunately, functions can not have numeric expressions as parameters.
%
Finally, we must define the initial and goal states for every instance. The initial state is simply stated with a \texttt{coloured} statement for each cell. However, the goal state is more complex to express if we do not have arithmetic or aggregate functions to count the number of cells coloured with \texttt{null}. In our instances we define the goal as follows. Let $g$ be the maximum allowed number of non-\texttt{null} cells in order to satisfy the goal state. We require that there exist $g$ different cells such that any other cell is \texttt{null}. E.g. requiring at most 2 non-\texttt{null} cells creates:

\begin{lstlisting}[escapechar=|, language=PDDL]
(:goal  ;; at most 2 cells are not null, i.e. g=2
  (exists (?x1 ?x2 ?y1 ?y2 - number) 
    (and (or (not (= ?x1 ?x2))
             (not (= ?y1 ?y2)))
       (forall (?x3 ?y3 - number) 
          (or ; Or is one of cell 1 or cell 2, or is null
            (and (= ?x1 ?x3) (= ?y1 ?y3)) 
            (and (= ?x2 ?x3) (= ?y2 ?y3))
            (coloured ?x3 ?y3 null))))))
\end{lstlisting}
The length of this goal is $\Theta(g^2)$, since the $g$ cells must be pair-wise different.
Again, this is simpler to state in a constraint language with, for example, an \texttt{atleast} constraint.

\section{Empirical Evaluation}\label{sec:results}
We improved our previous instance generator~\cite{espasa2022plotting} to avoid generating symmetric grids, ensuring more interesting instances. A new set of 522 instances was then created with a range of difficulties\footnote{Model and instances are available in \url{https://github.com/stacs-cp/Plotting-Journal}}. These use from 8 to 49 blocks, and from 2 to 6 colours.

We considered the best planners in the 2018 International Planning Competition. From all planners, 9 claimed to support the features required. Of those, 7 were based on the Fast Downward preprocessor and the rest crashed when given the instances. 
We therefore present results for only Fast Downward~\cite{fastdownward} 22.12 because the pre-processing for all planners based on Fast Downward is the same, and for the successfully pre-processed instances the search time was very small. The integrated Stone Soup portfolio showed that only the blind heuristic supported the features in the model. 

We also used a Planning as Satisfiability~\cite{kautzS92} approach and translating the problem to the constraint programming language Essence Prime. Then, \savilerow~\cite{savilerow} 1.9.1 was used to solve the problem with 
three different solvers: kissat 3.0.0, Chuffed 0.10.4, and OR-Tools 9.5. 
As a brief summary of the approach, a planning problem is encoded to a Boolean formula (or 
constraint satisfaction problem),
with the property that any model of this formula will correspond to a valid plan.
Since the length of a valid plan is not known a priori, we
encode the existence of a plan of $T$ steps with a formula $f(T)$. Then, the
method for finding the shortest plan consists in iteratively checking the
satisfiability of $f(T)$ for $T = 0,1,2, \dots$ until a satisfiable formula is
found.
%
%
%
We observed the lower bound on the number of blocks remaining in the grid is one less than the number of distinct colours in the initial grid. Combined with the fact that each shot should remove at least one block from the grid, for each instance we can consider a sequence of decision problems from $1$ up to $(\mathit{width}\times \mathit{height}) - \max(\mathit{goal}_b, \mathit{colours})$ steps, where $\mathit{goal}_b$ is the number of allowed blocks remaining in the goal states and $\mathit{colours}$ is the number of different colours in the grid. 
As expected, the hardest queries for each instance are the last unsatisfiable one (with the greatest queried time horizon among the unsatisfiable ones) and the first satisfiable query. 

Experiments were executed on a cluster of compute nodes with two 2.1 GHz 18-core Intel Xeon processors each. Each process was limited to 8GB of memory and 1 hour. 
In the table 
we compare instances solved within resource limits and the PAR2 score for each solver.
%
\begin{center}
\begin{tabular}{lrr}
\toprule
 Solver   & Instances Solved & PAR2 Score (s)  \\ \midrule
Chuffed  & \textbf{510}       & \textbf{162537}  \\
Kissat     & 508                & 176207      \\
OR-Tools & 488                & 329176           \\
FD       & 78                 & 3233535          \\ \bottomrule
\end{tabular}
\end{center}
The three CP solvers all perform reasonably well, being able to solve most instances within the given time.
In contrast, Fast Downward exhibits a significantly worse performance scaling, and also only solves the smallest instances, getting stuck during the grounding process.
When using the best constraint model, neither SAT, OR-Tools nor Chuffed ran out of memory. In contrast, out of the 522 instances, Fast Downward runs out of memory for 249 and times out for 195. Further, when Fast Downward exceeds resource bounds, it always does so during grounding.

\section{Conclusions}

As we have shown, classical planning can solve Plotting despite the highlighted PDDL limitations. Over the last few decades, the prevalent method of solving classical planning problems has been heuristic search. In such approaches, a grounded representation of the problem is generally needed to be able to then compute heuristic values that guide the search. The grounding component in most of the planners struggles when presented with the PDDL model.
More concretely, memory is exhausted due to the generation of large intermediate data structures, which is unavoidable if the expansion phase of grounding is performed before any pruning of the intermediate expressions. Plotting appears to be a \emph{hard-to-ground problem}~\cite{powerlifted}, and we therefore intend to investigate models that are easier to ground.

Problems with grounding could be mitigated by using a more expressive language, like those proposed by~\citet{pmt} or~\citet{fstrips2}, allowing more concise and efficient  problem representation. However, one may also need to deal with the grounding of this richer language to apply similar solving methods.
\citet{powerlifted} adapt the heuristic search framework to allow search in the non-grounded representation of the problem. Yet, those approaches are still too limited in their expressivity to be able to reason with essential constructs needed for Plotting, such as conditional effects and quantifiers.
Improvements to grounding, such as partial grounding~\cite{partialg} or pruning-during-grounding, may enable automated planning systems to leapfrog CP and SAT solvers for this hard-to-ground problem.




\section*{Acknowledgements}
Ian Miguel is funded by the EPSRC grant {EP/V027182/1}, Mateu Villaret is funded by MCIN/AEI/10.13039/501100011033 grant {PID2021-122274OB-I00} and by ERDF -- \textit{A way of making Europe}. Peter Nightingale is funded by EPSRC grant {EP/W001977/1}.
\bibliography{ref}

\end{document}